\documentclass[letterpaper,journal]{IEEEtran}
\usepackage{amsmath,amsfonts}
\usepackage{array}

\usepackage{textcomp}
\usepackage{xcolor}
\usepackage{stfloats}
\usepackage{url}
\usepackage{verbatim}
\usepackage{graphicx}
\usepackage{cite}
\usepackage[capitalize]{cleveref}
\usepackage{booktabs}
\usepackage{algorithm}
\usepackage{algorithmic}
\usepackage[caption=false,font=footnotesize]{subfig}
\usepackage{multirow} 
\usepackage{threeparttable}
\crefname{figure}{Fig.}{Figs.}
\Crefname{figure}{Figure}{Figures}

\begin{document}

\title{Bridging the Modality Gap in Roadside LiDAR: A Training-Free Vision-Language Model Framework for Vehicle Classification}

% TODO: add macros for easier formatting of \author.
\author{ Yiqiao Li*, Bo Shang, Jie Wei
\thanks{This research was supported by the U.S. Department of Transportation’s University Transportation Center—National Center for Understanding Future Travel Behavior and Demand (TBD), and the Research Foundation of the City University of New York (CUNY). The authors gratefully acknowledge this support. The authors would also like to express their gratitude to Dr. Andre Tok and Prof. Stephen Ritchie from the University of California, Irvine, for their generous support in sharing the data. The contents of this paper reflect the views of the authors and do not necessarily represent the official policies or positions of the sponsoring organizations. This paper does not constitute a standard, specification, or regulation.\\
Yiqiao Li (yli4@ccny.cuny.edu), Corresponding Author, Department of Civil Engineering, City College of New York \\
Bo Shang, Department of Civil Engineering, City College of New York  \\
Jie Wei, Department of Computer Science, City College of New York
}}

\maketitle

\begin{abstract}
Fine-grained truck classification is critical for intelligent transportation systems (ITS), yet current LiDAR-based methods face scalability challenges due to their reliance on supervised deep learning and labor-intensive manual annotation. Vision-Language Models (VLMs) offer promising few-shot generalization, but their application to roadside LiDAR is limited by a modality gap between sparse 3D point clouds and dense 2D imagery. We propose a framework that bridges this gap by adapting off-the-shelf VLMs for fine-grained truck classification without parameter fine-tuning. Our new depth-aware image generation pipeline 
applies noise
removal, spatial and temporal registration, orientation rectification, morphological operations, and anisotropic smoothing to transform sparse, occluded LiDAR scans into depth-encoded 2D visual proxies. Validated on a real-world dataset of 20 vehicle classes, our approach achieves competitive classification accuracy with as few as 16–30 examples per class, offering a scalable alternative to data-intensive supervised baselines. We further observe a “Semantic Anchor” effect: text-based guidance regularizes performance in ultra-low-shot regimes ($k<4$), but degrades accuracy in more-shot settings due to semantic mismatch. Furthermore, we demonstrate the efficacy of this framework as a ``Cold Start" strategy, using VLM-generated labels to bootstrap lightweight supervised models. Notably, the few-shot VLM-based model achieves over correct classification rate of 75 percent for specific drayage categories (20ft, 40ft, and 53ft containers) entirely without the costly training or fine-tuning, significantly reducing the intensive demands of initial manual labeling, thus achieving a method of practical use in ITS applications.
\end{abstract}

%%%%%%%%% BODY TEXT
\section{Introduction}
\label{sec:intro}

Heavy-duty trucks, while constituting a minor fraction of total traffic volume, exert a disproportionate impact on transportation infrastructure and safety. They account for the vast majority of pavement fatigue~\cite{aashto1993guide}, critical bridge loading events~\cite{nowak1993effect}, and severe collisions due to their substantial mass and limited maneuverability~\cite{jansen2022caught}. Furthermore, the heterogeneity of commercial fleets, ranging from bobtails to multi-trailer configurations, complicates emissions modeling~\cite{lu2024moves} and tolling operations. Consequently, accurate and fine-grained identification of truck body configurations is an important requirement for infrastructure management and intelligent transportation systems (ITS).

To meet this requirement, researchers have increasingly turned from passive video surveillance—which suffers from performance degradation under low-light and adverse weather—to active sensing technologies \cite{sathyamoorthy2026ensemble}. Infrastructure-based LiDAR (Light Detection and Ranging) has become a preferred choice for 24/7 persistent monitoring, offering geometric data that is largely independent of ambient illumination~\cite{zhao2019detection, li2023lidar,li2025adaptive}. However, the operational deployment of LiDAR-based classification faces a significant bottleneck: reliance on supervised deep learning. State-of-the-art frameworks typically require large, manually annotated datasets to achieve high accuracy. This dependency creates a ``scalability trap'': each new deployment site, with its unique sensor geometry and occlusion patterns, may require costly and time-consuming re-annotation and model retraining~\cite{li2021truck}.

The emergence of Vision-Language Models (VLMs) such as CLIP~\cite{radford2021learning} offers a potential alternative. Trained on web-scale image-text pairs, VLMs exhibit zero- and few-shot generalization capabilities that may allow systems to recognize novel vehicle classes without extensive retraining. However, applying VLMs to roadside LiDAR introduces two gaps:
\begin{enumerate}
    \item \textbf{The Modality Gap:} VLMs' visual encoders are optimized for dense, texture-rich RGB images. Raw LiDAR point clouds are sparse and textureless, yielding low-fidelity projections that may not align well with the model's pre-trained feature expectations.
    \item \textbf{The Reality-to-Sim Gap:} Most 3D-VLMs are trained on complete ``watertight'' CAD models. Roadside sensors, in contrast, capture only a ``half-shell'', or even a small portion, of the vehicle due to self-occlusion.
\end{enumerate}
Furthermore, while the language component of VLMs is powerful, natural language's inherent ambiguity \cite{khan2024consistency} raises questions about its utility in specialized engineering domains. It remains unclear whether natural language-based descriptions of technical vehicle types (e.g., ``20ft Container Truck'') can effectively identify the model's latent embedding vector space.

\begin{figure*}[htbp]
   \centering
    \includegraphics[width=0.95\linewidth]{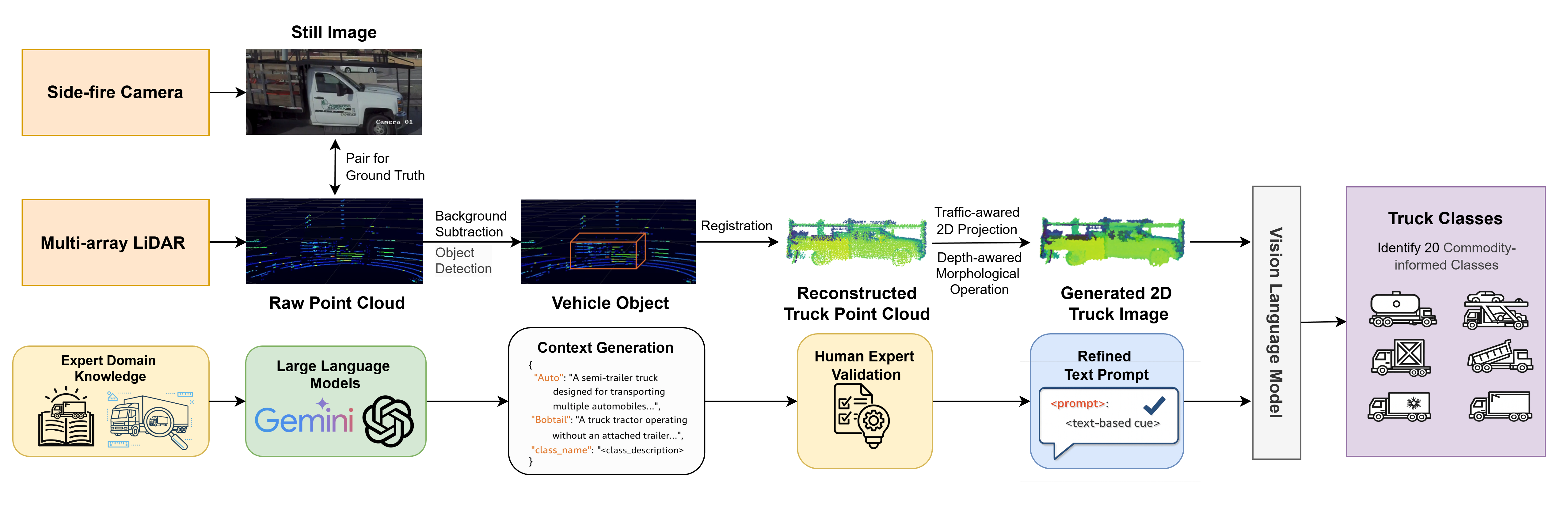}
    \caption{Flowchart of the proposed framework.}
    \label{fig:framework}
\end{figure*}

To address these gaps, we propose a new framework that adapts general-purpose VLMs for fine-grained truck classification using sparse roadside LiDAR data. Moving beyond simple projection techniques, we introduce a reconstruction-based pipeline that converts sparse data into visual proxies suitable for VLM encoders. The proposed framework is outlined diagrammatically in Figure \ref{fig:framework}.

The main contributions of this work are as follows:
\begin{enumerate}
    \item \textbf{A Novel Depth-Aware Image Generation Pipeline:} We propose an image generation method that registers temporal point cloud sequences and applies outlier removal, mathematical morphological operators, and bilateral filtering to mitigate sensor sparsity. This pipeline effectively transforms half-shell LiDAR scans into depth-encoded 2D images that register better with the visual distributions expected by pre-trained VLM encoders.
    \item \textbf{Data-Efficient Few-Shot Adaptation:} We validate our framework on a real-world dataset of 20 fine-grained vehicle classes. The proposed method achieves competitive classification accuracy with as few as 16--30 examples per class, offering a scalable alternative to data-intensive supervised baselines.
    \item \textbf{Analysis of the Semantic Gap:} We provide a systematic analysis of the trade-off between visual and textual features in ITS applications. Our experiments reveal a \textit{Semantic Anchor} effect: while text embeddings generally degrade performance due to domain mismatch, they provide useful regularization in ultra-low-shot regimes ($k < 4$). This suggests that visual-centric adaptation tends to be preferable for specialized industrial tasks as data increases.
    \item \textbf{Comprehensive Comparison with Supervised Adaptation:} We present a systematic comparison between our training-free few-shot method, conventional supervised adaptation, and fully supervised learning approaches. Our results show that large supervised models (e.g., ViT-B/16) perform poorly with limited data, while smaller supervised models achieve moderate accuracy with 30-shot settings but require extensive adaptation time. In contrast, the VLM-based method—especially our few-shot approach—requires no training and attains competitive accuracy with only 16–30 examples per class, though inference is slower than supervised models. We recommend using the VLM-based method as a ``Cold Start" strategy to generate labels for lightweight supervised models, enabling efficient real-time deployment. Furthermore, our attention heatmap analysis reveals that larger VLMs focus on fine-grained features of critical importance for truck classification, such as trailer gaps and axle counts, whereas smaller VLMs exhibit more diffuse attention, including background regions. Full image generation enhances the model's focus on vehicle objects and salient features, improving classification performance.
\end{enumerate}

\section{Related Work}
\label{sec:related}

\subsection{LiDAR-based Vehicle Classification}
While vision-based systems utilizing cameras have traditionally been the standard for vehicle classification, their efficacy is strictly constrained by ambient lighting conditions. Camera performance degrades significantly during nighttime operations, which hinders consistent truck monitoring. In contrast, LiDAR functions as an active sensor, remaining minimally affected by illumination changes. This inherent environmental robustness has prompted a shift toward LiDAR-based frameworks for persistent, 24/7 traffic monitoring.

Early investigations into LiDAR-based classification in the 2000s utilized overhead mountings to scan the roadway cross-section \cite{abdelbaki2002, hussain2005laser}.  While these systems effectively captured detailed vehicle features \cite{sandhawalia2013}, their widespread deployment was impeded by significant infrastructure constraints and the prohibitive costs of overhead gantry installations. Consequently, the field shifted focus toward more accessible roadside (side-fire) setups.

Early roadside approaches, such as those by Lee and Coifman \cite{lee2012side}, employed vertically mounted laser scanners to extract high-level physical clusters. However, vertical scanning suffers from an extremely narrow Field-of-View (FOV); when a vehicle occupies the outermost lane, it completely occludes traffic in inner lanes, causing severe information loss \cite{asborno2019truck}.  Subsequent efforts using low-cost single-beam LiDAR proved cost-efficient but failed to provide the geometric resolution required for fine-grained classification.

To reconcile the trade-off between resolution and FOV, researchers explored temporally stitching vertical scan frames \cite{vatani2019transfer, sahin2021methods}. However, these methods rely on a rigid constant-speed assumption, causing performance to deteriorate under real-world variable traffic conditions. Conversely, horizontally oriented sensors broaden the detection zone but capture only sparse point clouds. To mitigate this, recent frameworks \cite{li2021truck,li2023lidar,li2025lidar} utilize computational reconstruction to generate dense vehicle representations from sparse inputs. While these supervised learning methods achieve high accuracy, they necessitate labor-intensive data annotation and large-scale training sets. This reliance on costly manual labeling limits their generality and scalability, motivating the need for data-efficient, few-shot alternatives.

\subsection{Vision-Language Models and the Domain Gap}
The advent of deep learning introduced architectures such as PointNet and PointNet++ \cite{qi2017pointnet,qi2017pointnet++}, which operate directly on 3D data. While robust, these models suffer from high sample complexity, requiring massive domain-specific datasets that are expensive to annotate in industrial contexts under practical scenarios. 

Concurrently, VLMs such as CLIP \cite{radford2021learning} and EVA \cite{fang2023eva} have revolutionized recognition by learning joint representations from web-scale image-text pairs. Their few-shot capabilities make them ideal for data-scarce domains. However, applying them to LiDAR data introduces two major gaps:
\subsubsection{The Modality Gap}
VLMs are optimized for dense 2D images. Applying them to sparse 3D point clouds requires bridging the ``modality gap". Common strategies involve projecting 3D data into 2D views \cite{zhang2022pointclip}. However, raw LiDAR projections often lack the visual fidelity required for pre-trained image encoders to extract meaningful features.

\subsubsection{The Reality-to-Sim Gap (The "Half-Shell" Problem)}
State-of-the-art 3D-text alignment models, such as OpenShape \cite{liu2023openshapescaling3dshape} and ULIP \cite{xue2023ulip}, are predominantly trained on ``watertight" CAD models (e.g., Objaverse). In roadside applications, LiDAR sensors capture only a ``half-shell" of the vehicle due to self-occlusion and sensor perspective.  This discrepancy creates a ``reality-to-sim" gap where models trained on complete geometry may fail to identify critical features in partial real-world scans. Furthermore, single-frame projection methods often lack the density to reveal fine-grained details—such as axle counts or trailer gaps—essential for regulatory classification.

\subsection{Synthesis and Research Gap}
Recent advancements have sought to strengthen 3D-VLM interfaces using high-fidelity proxies like Gaussian Splatting \cite{li2024gsclipgaussiansplattingcontrastive} or promptable segmentation \cite{zhou2024pointsampromptable3dsegmentation}. However, a gap remains: existing state-of-the-art models typically rely on either complete 3D geometry or extensive supervised fine-tuning.

To address these challenges, we introduce a novel framework that adapts general-purpose VLMs for fine-grained truck classification using sparse roadside LiDAR data. By combining images generated by depth-aware morphological reconstruction from sparse 3D point clouds with domain-aware prompt engineering, our approach enables robust few-shot learning and operational scalability, even in data-scarce environments.
\section{Preliminary}
\subsection{Vision Language Models (VLMs)}
VLMs are multimodal systems that process and integrate both visual and textual data to generate text or image outputs. Leveraging large-scale pre-training, VLMs exhibit strong zero/few-shot performance on diverse tasks such as visual question answering, image captioning, and instruction-driven recognition~\cite{lin2024vila}. Advanced variants can also perform spatial reasoning, enabling them to detect and segment objects within an image when prompted~\cite{lu2024deepseek}. A VLM's specific capabilities are determined by its underlying architecture, training data, and image encoding strategy, leading to a wide range of performance across different applications.
\subsubsection{The Contrastive Language-Image Pre-training (CLIP)}
CLIP~\cite{radford2021learning} is a foundational vision-language model with strong zero/few-shot classification capabilities. CLIP is trained on a massive dataset of 400 million image-text pairs collected from the Internet. It consists of two primary components: a text encoder (a Transformer) and an image encoder, {\it e.g.}, a Vision Transformer (ViT)~\cite{dosovitskiy2020image}. During pre-training, CLIP learns a multimodal embedding space by jointly training these encoders to maximize the similarity of the embeddings of correct image-text pairs while minimizing the similarity of incorrect pairs under the {\it contrastive learning}~\cite{wei2021nida} framework.

Formally, given a batch of $N$ (image, text) pairs, let the image and text embeddings be denoted by $\mathbf{e}_i^I$ and $\mathbf{e}_i^T$ respectively, where $i=1, \dots, N$. The embeddings are L2-normalized. CLIP computes a matrix of cosine similarities $S \in \mathbb{R}^{N \times N}$, where $S_{ij} = (\mathbf{e}_i^I)^T \mathbf{e}_j^T$. The model is trained by minimizing a symmetric cross-entropy loss over the similarity scores. For the image-to-text direction, the contrastive loss is defined below:
\begin{equation}
    \mathcal{L}_{\text{I2T}} = -\frac{1}{N}\sum_{i=1}^{N} \log \frac{\exp(S_{ii}/\tau)}{\sum_{j=1}^{N}\exp(S_{ij}/\tau)},
\end{equation}
where $\tau$ is a learned temperature parameter. A symmetric loss, $\mathcal{L}_{\text{T2I}}$, is computed for the text-to-image direction, and the total CLIP loss is the average of the two: $\mathcal{L}_{\text{CLIP}} = (\mathcal{L}_{\text{I2T}} + \mathcal{L}_{\text{T2I}})/2$. This objective effectively pulls the embeddings of corresponding pairs together in the feature space while pushing non-corresponding pairs apart.

For a zero-shot classification task, CLIP uses the text encoder to embed natural language descriptions of the target classes, e.g., ``a photo of a bobtail truck". Given an image, the image encoder produces a visual embedding. The model then predicts the class corresponding to the text description that has the highest cosine similarity to the image embedding. This allows CLIP to classify images into categories for which it has not been explicitly trained. In this study, we adapt this model to the few-shot classification of our point-cloud-projected images, leveraging two common variants of its ViT image encoder: ViT-B/32 and ViT-L/14.
  
\subsubsection{Exploring the Limits of Visual Representation at Scale (EVA)}
EVA is a vision-centric foundation model designed to learn powerful representations from publicly accessible large-scale data~\cite{fang2023eva}. It uses a vanilla Vision Transformer (ViT) architecture and is pre-trained using a Masked Image Modeling (MIM) objective, where it learns to reconstruct masked image-text aligned vision features, i.e., CLIP features,  conditioned on the visible image patches. This pretraining strategy allows EVA to scale effectively to more than one billion parameters and has demonstrated state-of-the-art performance on a wide range of downstream vision tasks. For our comparative analysis, we use the EVA-L variant to assess the performance of our pipeline against another powerful, publicly available vision model.

\section{Methodology}
The proposed framework integrates temporal point cloud reconstruction with VLM adaptation to address the challenges of sparse LiDAR data. The pipeline consists of two primary modules: (1) \textit{Depth-Aware Image Generation}, which transforms raw 3D point sequences into high-fidelity 2D images as visual proxies, and (2) \textit{Visual-Semantic Prompt Engineering}, which adapts pre-trained VLMs for fine-grained vehicle classification via few-shot in-context learning.

\subsection{Depth-Aware Image Generation}
The initial data processing follows the pipeline established in our prior work~\cite{li2021truck, li2023lidar}. Raw LiDAR data is first processed to segment individual vehicles from the background using a combination of background subtraction and DBSCAN clustering~\cite{schubert2017dbscan}. These segmented vehicles are then tracked across consecutive frames using the SORT algorithm~\cite{bewley2016simple}.

\subsubsection{Statistical Outlier Removal on 3D Points}
To refine the spatial data, the point clouds went through a two-stage preprocessing pipeline applied to each single-frame object: \textit{voxel downsampling} followed by \textit{statistical outlier removal (SOR)}~\cite{zhou2018open3d}.

\paragraph{1. Voxel Downsampling:} To reduce computational overhead while preserving the underlying geometric structure, the 3D space is partitioned into regular cubic voxels. Each voxel is represented by a single centroid point. In this study, a voxel size of $0.05$~m was utilized.

\paragraph{2. Outlier Removal:} Following downsampling, an SOR filter is applied to eliminate noise. For each point $p_i$ in the downsampled point cloud, the algorithm identifies its $k$ nearest neighbors to calculate local density metrics. Let the set of $k$ nearest neighbors for point $p_i$ be denoted as:
\begin{equation}
\mathcal{N}_i = \{q_{i1}, q_{i2}, \dots, q_{ik}\}
\end{equation}
where $k = 20$ in this study.

\textbf{Local Distance Statistics:} We compute the mean Euclidean distance from $p_i$ to its neighbors:
\begin{equation}
\mu_i = \frac{1}{k} \sum_{j=1}^{k} \|p_i - q_{ij}\|
\end{equation}
where $\|p_i - q_{ij}\|$ represents the Euclidean distance:
\begin{equation}
\|p_i - q_{ij}\| = \sqrt{(x_i - x_{ij})^2 + (y_i - y_{ij})^2 + (z_i - z_{ij})^2}
\end{equation}

\textbf{Global Statistics:} The global mean $\mu_g$ and standard deviation $\sigma_g$ of the local mean distances across the entire point cloud of $n$ points are computed as:
\begin{align}
\mu_{g} &= \frac{1}{n} \sum_{i=1}^{n} \mu_i \\
\sigma_{g} &= \sqrt{\frac{1}{n-1} \sum_{i=1}^{n} (\mu_i - \mu_{g})^2}
\end{align}

\textbf{Outlier Detection:} A point $p_i$ is classified as an outlier and removed if its local mean distance significantly exceeds the global average:
\begin{equation}
\mu_i > \mu_{g} + \alpha \cdot \sigma_{g}
\end{equation}
where $\alpha$ is a sensitivity threshold. We set $\alpha = 1.0$ to effectively remove noise while retaining vehicle structure.

\subsubsection{Temporal Point Cloud Reconstruction}
\begin{figure}[!t]
\centering
\subfloat[Single Frame\label{fig:single_frame}]{
  \includegraphics[width=0.45\linewidth]{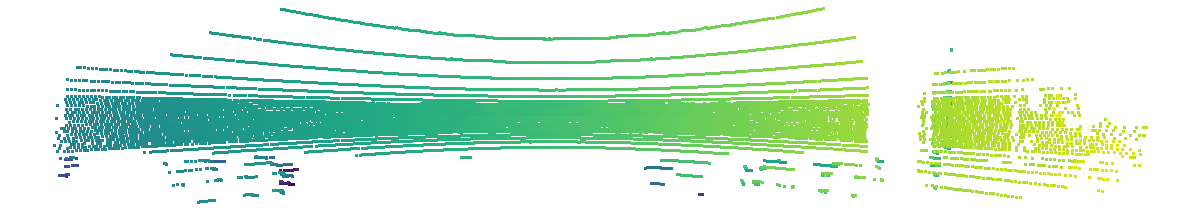}
}
\hfill
\subfloat[Reconstructed Frame\label{fig:reconstruct}]{
  \includegraphics[width=0.45\linewidth]{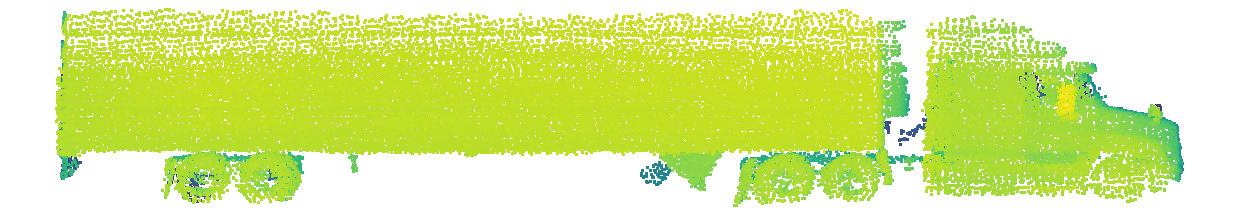}
}
\caption{Comparison of single frame and reconstructed frame (color represents depth information).}
\label{fig:frame_comparison}
\end{figure}

A single LiDAR frame provides a sparse point cloud that is ill-suited for VLMs pre-trained on dense 2D images. To bridge this modality gap, we register and fuse the filtered point clouds from consecutive frames of each tracked vehicle. This process aligns the frames using a probabilistic registration method~\cite{gao2019filterreg} to reconstruct a single, dense 3D representation of the vehicle. As illustrated in Figure~\ref{fig:frame_comparison}, the resulting aggregated point cloud provides a clearer, geometry-rich depiction of the vehicle, forming a robust basis for the ensuing 2D image rendering.

\subsubsection{Canonical Orientation and 2D Projection}
After reconstruction, the vehicle's travel direction is estimated from the transformation matrix obtained during multi-frame registration. This direction is represented by a motion vector $\mathbf{d}$. To determine a canonical orientation, the point cloud is aligned within a unique Cartesian coordinate system using three orthonormal axes derived from the vehicle motion and the ground plane.

Let $\mathbf{g}$ denote the \textbf{unit} ground normal vector. We define the canonical vertical axis (Vehicle Height) as:
\begin{equation}
\hat{\mathbf{z}} = \mathbf{g}.
\end{equation}

The vehicle motion $\mathbf{d}$ is projected onto the plane orthogonal to $\hat{\mathbf{z}}$ to ensure it lies on the ground plane:
\begin{equation}
\mathbf{d}_{\perp} = \mathbf{d} - (\mathbf{d}^{\top}\hat{\mathbf{z}})\,\hat{\mathbf{z}}.
\end{equation}

Since the vehicle is defined to travel along the \textbf{negative} $Y$-axis, we define the canonical longitudinal axis $\hat{\mathbf{y}}$ as the negative normalized motion vector:
\begin{equation}
\hat{\mathbf{y}} = - \frac{\mathbf{d}_{\perp}}{\|\mathbf{d}_{\perp}\|}.
\end{equation}

The lateral depth axis $\hat{\mathbf{x}}$ is then defined to complete the right-handed orthonormal basis ($\hat{\mathbf{x}} = \hat{\mathbf{y}} \times \hat{\mathbf{z}}$):
\begin{equation}
\hat{\mathbf{x}} = \hat{\mathbf{y}} \times \hat{\mathbf{z}}.
\end{equation}

Together, $\{\hat{\mathbf{x}}, \hat{\mathbf{y}}, \hat{\mathbf{z}}\}$ form the canonical basis:
\begin{itemize}
    \item $X$-axis ($\hat{\mathbf{x}}$): Lateral (Depth information)
    \item $Y$-axis ($\hat{\mathbf{y}}$): Longitudinal (Opposite to Travel Direction)
    \item $Z$-axis ($\hat{\mathbf{z}}$): Vertical (Vehicle Height)
\end{itemize}

This alignment procedure ensures that the projection onto the $YZ$ plane yields a consistent \textit{side view} of the vehicle.

Each 3D point cloud $\mathcal{C}$ is rotated into this canonical coordinate system. The corresponding 2D projection image $I$ is generated by mapping the longitudinal ($y$) and vertical ($z$) coordinates to the image grid and capturing the minimum value along the depth axis ($x$). For a pixel location $(y, z)$ in the image grid:
\begin{equation}
I(y, z) = \min \left\{ p_x \;\middle|\; p \in \mathcal{C},\; |p_y - y| < \delta,\; |p_z - z| < \delta \right\}
\label{eq:3d22d}
\end{equation}
where $(p_x, p_y, p_z)$ denote the transformed coordinates of a point $p$ in the canonical frame, and $\delta$ defines the grid resolution for spatial aggregation.

\subsubsection{Depth-Aware Image Smoothing}
To encode more information into the converted image $I$ for VLM input, we apply a two-stage filtering process.

\paragraph{Stage 1: Morphological Reconstruction.} Morphological opening (erosion followed by dilation)~\cite{najman2013mathematical} is applied to perform depth-aware smoothing and mitigate sparsity effects. Given input image $I$, the operation is defined as:
\begin{equation}
    I_{\text{opened}} = (I \ominus K) \oplus K
\end{equation}
where $\ominus$ and $\oplus$ denote erosion and dilation, respectively, and $K$ is an elliptical structuring element of size $s \times s$. This step effectively removes high-frequency noise artifacts introduced by the projection in Eq.~(\ref{eq:3d22d}).

\paragraph{Stage 2: Bilateral Filtering.} A bilateral filter is then applied to suppress residual noise while preserving sharp geometric boundaries. For each pixel location $p=(x,y)$, the filtered intensity is computed as:
\begin{equation}
I_{\text{final}}(p) = \frac{\sum_{q \in \Omega_p} I_{\text{opened}}(q)\, w(p,q)} {\sum_{q \in \Omega_p} w(p,q)}
\end{equation}
where the bilateral weight $w(p,q)$ is defined as:
\begin{equation}
w(p,q) = \exp \left( -\frac{\|p - q\|^2}{2\sigma_{sp}^2} -\frac{\left| I_{\text{opened}}(p) - I_{\text{opened}}(q) \right|^2}{2\sigma_c^2} \right)
\end{equation}
Here, $\Omega_p$ denotes the spatial neighborhood of pixel $p$, $\sigma_{sp}$ controls the spatial smoothing extent, and $\sigma_c$ governs sensitivity to intensity (depth) differences.

In summary, the complete depth-aware image generation algorithm is outlined in Algorithm~\ref{alg:depth_aware}.

\begin{algorithm}[!t]
\caption{Depth-Aware Image Generation}
\label{alg:depth_aware}
\begin{algorithmic}[1]
\REQUIRE Inputs: Raw 3D point cloud sequence $\mathcal{S}_{raw} = \{F_1, F_2, \dots, F_T\}$, Voxel size $\nu$, Neighbors $k$, Threshold $\alpha$, Structuring element $K$.
\ENSURE Output: Processed depth-aware image $I_{final}$.

\STATE $\mathcal{S}_{filtered} \leftarrow \emptyset$

\STATE \textbf{Step 1: Frame-wise Outlier Removal}
\FOR{each frame $F_t$ in $\mathcal{S}_{raw}$}
    \STATE \textit{// Voxel Downsampling}
    \STATE $F'_t \leftarrow \text{Downsample}(F_t, \nu)$
    
    \STATE \textit{// Statistical Outlier Removal (SOR)}
    \STATE Compute mean neighbor distance $\mu_i$ for all points in $F'_t$.
    \STATE Compute global stats $\mu_g, \sigma_g$ for $F'_t$.
    \STATE $F''_t \leftarrow \{p \in F'_t \mid \mu_p \leq \mu_g + \alpha \cdot \sigma_g\}$
    
    \STATE Add $F''_t$ to $\mathcal{S}_{filtered}$.
\ENDFOR

\STATE \textbf{Step 2: Temporal Point Cloud Reconstruction}
\STATE \textit{// Fuse all filtered frames into one dense cloud}
\STATE $\mathcal{C} \leftarrow \text{ProbabilisticRegistration}(\mathcal{S}_{filtered})$

\STATE \textbf{Step 3: Canonical Orientation and 2D Projection}
\STATE Compute motion vector $\mathbf{d}$ and ground normal $\mathbf{g}$.
\STATE Orthogonalize $\mathbf{d}$: $\mathbf{d}_{\perp} \leftarrow \mathbf{d} - (\mathbf{d}^\top \mathbf{g})\mathbf{g}$.
\STATE Normalize basis: $\hat{\mathbf{d}} \leftarrow \mathbf{d}_{\perp} / \|\mathbf{d}_{\perp}\|$, $\hat{\mathbf{g}} \leftarrow \mathbf{g} / \|\mathbf{g}\|$.
\STATE Compute lateral vector: $\hat{\mathbf{v}} \leftarrow \hat{\mathbf{d}} \times \hat{\mathbf{g}}$.
\STATE Rotate $\mathcal{C}$ to align with canonical basis $\{\hat{\mathbf{d}}, \hat{\mathbf{g}}, \hat{\mathbf{v}}\}$.

\STATE Initialize image grid $I_{raw}$.
\FOR{each pixel $(x,y)$}
    \STATE $I_{raw}(x, y) \leftarrow \min \{p_z \mid |p_x-x|< \delta, |p_y-y| < \delta, p \in \mathcal{C}\}$
\ENDFOR

\STATE \textbf{Step 4: Depth-Aware Smoothing}
\STATE $I_{opened} \leftarrow (I_{raw} \ominus K) \oplus K$
\STATE $I_{final} \leftarrow \text{BilateralFilter}(I_{opened}, \sigma_{sp}, \sigma_c)$

\RETURN $I_{final}$
\end{algorithmic}
\end{algorithm}

Our algorithm encodes vehicle structure from 3D points into 2D images through spatial-temporal registration, orientation rectification, and anisotropic smoothing. By mitigating noise from both the sparse sensor data and the projection process, we yield a clean representation suitable for VLM encoders, effectively leveraging visual features learned from large-scale 2D pre-training.

\subsection{Visual-Semantic Prompt Engineering}
While VLMs show strong zero-shot abilities for general content, they may struggle with complex tasks in zero-shot settings—particularly when processing point-cloud projected images that differ from their pre-training distribution. To address this, we provide VLMs with few-shot demonstrations following In-Context Learning (ICL) approaches~\cite{brown2020language}. The demonstrations condition the model on the target task and help improve classification accuracy. The description of each vehicle class used in the prompt is presented in Table \ref{tab:vehicle_classes}.

We adopt a domain-aware few-shot prompting strategy based on~\cite{brown2020language} for vehicle point-cloud projected image classification. The prompts comprise three components: task instructions, class definitions, and support set examples. The prompt design is illustrated in Figure \ref{fig:prompt}.

To quantify the impact of semantic information, we modeled the classification as a prototype fusion, {\it i.e.,} a weighted linear combination of visual and textual embeddings. The fused prototype $p_c^{\text{fused}}$ for class $c$ is defined as:

\begin{equation}
\label{eq:fusion}
p_c^{\text{fused}} = (1-w) \cdot p_c^{\text{visual}} + w \cdot e_c^{\text{text}}
\end{equation}

\noindent where $p_c^{\text{visual}}$ is the mean embedding of the visual support set, $e_c^{\text{text}}$ is the text embedding from the language encoder, and $w \in [0, 1]$ represents the scalar text weight. 

\begin{figure}[!t]
\centering
\footnotesize
\fbox{%
\begin{minipage}{0.9\columnwidth}
\textbf{Domain-Aware Few-Shot Prompt Design Framework}

\medskip
\textbf{Task Instruction:} 
\textit{Classify a vehicle from its 3D LiDAR point cloud projection by leveraging domain-specific visual and semantic cues.}

\medskip
\textbf{Class Descriptions:}
\begin{itemize}
    \item \textbf{53ft Container:} \textit{A semi-trailer with a long, rectangular container extending over the rear axles.}
    \item \textbf{Bobtail:} \textit{A truck tractor operating without a trailer (only the cab and drive wheels).}
    \item \textbf{Tank (Semi):} \textit{A semi-trailer with a cylindrical tank designed for liquid or gas transport.}
\end{itemize}

\medskip
\textbf{Few-Shot Examples:}
\begin{itemize}
    \item Example 1: \textit{[53ft Container Image]} $\rightarrow$ \textbf{53ft Container}
    \item Example 2: \textit{[Bobtail Image]} $\rightarrow$ \textbf{Bobtail}
    \item Example 3: \textit{[Tank Semi Image]} $\rightarrow$ \textbf{Tank (Semi)}
\end{itemize}

\medskip
\textbf{Query:} \textit{[New Image]} $\rightarrow$ \textbf{?}

\medskip
\textbf{Prompt Templates:} 
{\textit{"Class Name"} : \textit{"Corresponding Class Description presented in Table \ref{tab:vehicle_classes}"}}
\end{minipage}%
}
\caption{Illustration of the proposed domain-aware few-shot prompt design for vision–language model (VLM)-based truck classification.}
\label{fig:prompt}
\end{figure}

\begin{table}[t]
\small
\centering
\caption{Vehicle Class Descriptions}
\label{tab:vehicle_classes}
\setlength{\tabcolsep}{4pt}
\renewcommand{\arraystretch}{1.05}
\begin{tabular}{@{}
>{\raggedright\arraybackslash}p{0.23\columnwidth}
>{\raggedright\arraybackslash}p{0.72\columnwidth}
@{}}
\toprule
\textbf{Class} & \textbf{Description} \\
\midrule
Auto & Semi-trailer truck for transporting multiple automobiles using ramps or multi-level decks. \\
Bobtail & Truck tractor operating without an attached trailer, typically for repositioning or short trips. \\
Enclosed Van (SU) & Single-unit truck with an enclosed cargo box for local delivery or secure freight. \\
Enclosed Van (Semi) & Semi-trailer truck with a fully enclosed van (dry van) trailer, sometimes with skirts or refrigeration units. \\
End Dump (Semi) & Semi-trailer with an open rectangular box that unloads bulk material by tipping backward hydraulically. \\
Light Duty & Small single-unit truck designed for light loads and maneuverability in urban settings. \\
Low Boy Platform & Semi-trailer flatbed with an extra-low deck for oversized or heavy equipment. \\
Low Loading & Single-unit truck with a low cargo bed to facilitate manual or forklift loading. \\
Passenger Vehicle & Vehicle designed primarily for passenger transport (e.g., car, SUV, minivan). \\
Pickup--Utility--Service & Pickup or utility truck configured for maintenance or service tasks with tool racks or boxes. \\
Pickup--Utility--Service w/ Trailer & Pickup or utility vehicle towing a small trailer for additional equipment or materials. \\
Platform (Semi) & Semi-trailer flatbed with an open platform for large or irregular cargo. \\
Platform (SU) & Single-unit flatbed truck for transporting pallets, equipment, or construction materials. \\
Platform Trailer & Single-unit platform truck towing a small trailer to extend cargo capacity. \\
Stake Body (SU) & Single-unit truck with stakes or removable side panels for bulk or irregular loads. \\
Tank (Semi) & Semi-trailer with a cylindrical tank for transporting liquids or gases. \\
Tank Tank & Truck or trailer equipped with a cylindrical tank body for liquid or flowable materials. \\
20\,ft Container & Semi-trailer hauling a standard 20-foot intermodal container for dense or heavy cargo. \\
40\,ft Container & Semi-trailer hauling a standard 40-foot intermodal container widely used in freight transport. \\
53\,ft Container & Semi-trailer hauling a 53-foot domestic container commonly used in U.S. freight operations. \\
\bottomrule
\end{tabular}
\end{table}

\section{Experimental Setup}
\subsection{Dataset}
The proposed method was evaluated on a dataset collected at the San Onofre Truck Scale on the I-5S freeway in Southern California~\cite{li2023lidar}. The data, captured over three weeks in 2019, includes a diverse range of truck traffic under both free-flow and congested conditions. A Velodyne VLP-32c LiDAR sensor was mounted on a 2\,m elevated platform to capture vehicle point clouds, synchronized with a video camera for ground truth verification. The dataset comprises 20 fine-grained vehicle classes manually labeled via camera-based visual verification for this study.

\begin{table}[t]
\centering
\caption{Training setup and hyperparameters for each experiment. LR: learning rate; BS: batch size; WD: weight decay.}
\label{tab:hyperparameters}
\small
\setlength{\tabcolsep}{3pt}
\renewcommand{\arraystretch}{1.2}
\begin{tabular}{l p{5.2cm}}
\toprule
\textbf{Method} & \textbf{Training Setup \& Hyperparameters} \\
\midrule
Ours (Few-Shot) & Training-free. CLIP ViT-L/14; class prototype = mean of support-set visual embeddings. \\
\midrule
Ours (Linear Probe) & Frozen CLIP ViT-L/14; linear head only. 80 epochs, LR=$10^{-3}$, BS=32, WD=$10^{-4}$, AdamW. \\
\midrule
ViT-B/16 (Scratch) & Full fine-tuning (backbone unfrozen). 20 epochs, LR=$5\times10^{-5}$, BS=32, WD=$10^{-4}$, AdamW. \\
\midrule
PointNet & 20 epochs, LR=$10^{-3}$, BS=32, WD=$10^{-4}$, AdamW. 1024 pts/sample, unit-sphere normalization. \\
\bottomrule
\end{tabular}
\end{table}

\subsection{Implementation Details}
For the few-shot classification, two variants of the CLIP Vision Transformer (ViT) were employed as image encoders: ViT-B/32 (standard) and ViT-L/14 (large). The large-scale EVA02-L-14 model was also included as a strong vision-only baseline. To simulate realistic few-shot constraints, $k \in \{1, 2, 4, 8, 16, 20, 30\}$ examples were randomly sampled per class to serve as the support set. Performance was evaluated on the remaining examples. To ensure statistical validity and mitigate selection bias, each configuration was evaluated over 10 independent experimental rounds. Performance is reported as the mean F1-score with standard deviation ($\sigma$). All experiments were conducted on a single NVIDIA RTX 4090 GPU.

\subsection{Training Setup and Hyperparameters}
\label{sec:training_setup}
Table \ref{tab:hyperparameters} documents the training configuration and hyperparameters for each experiment. The few-shot learning method (Ours, Few-Shot Learning) is training-free: it computes class prototypes as the mean of CLIP ViT-L/14 visual embeddings over the support set, with no optimization. The Linear Probe trains only a linear classification head on frozen CLIP features; we report results at 80 epochs, which yielded optimal F1 (0.649) in our sensitivity analysis. The ViT-B/16 baseline performs full fine-tuning (backbone unfrozen) from scratch with a lower learning rate to avoid catastrophic forgetting. All supervised methods use cross-entropy loss and 10 repeated runs with different random seeds for statistical reporting.

\begin{table*}[ht]
\centering
\begin{threeparttable}
\caption{Few-Shot Performance Comparison (Mean F1-Score)}
\label{tab:fewshot-results}
\setlength{\tabcolsep}{10pt}
\renewcommand{\arraystretch}{1.2}
\begin{tabular}{lcccccccc}
\hline
\textbf{Model} & \textbf{0 Shot} & \textbf{1 Shot} & \textbf{2 Shots} & \textbf{4 Shots} & \textbf{8 Shots} & \textbf{16 Shots} & \textbf{20 Shots} & \textbf{30 Shots} \\
\hline
EVA02-L-14 (Original)\tnote{1} & 0.08 & 0.24 & 0.30 & 0.34 & 0.39 & 0.43 & 0.43 & 0.44 \\
EVA02-L-14 (Generated)\tnote{2} & 0.08 & 0.25 & 0.31 & 0.37 & 0.40 & 0.44 & 0.45 & 0.47 \\
CLIP-B-32 (Original)\tnote{1}   & 0.05 & 0.29 & 0.37 & 0.48 & 0.54 & 0.59 & 0.60 & 0.61 \\
CLIP-B-32 (Generated)\tnote{2}   & 0.05 & 0.30 & 0.37 & 0.45 & 0.52 & 0.55 & 0.55 & 0.56 \\
CLIP-L-14 (Original)\tnote{1}   & 0.02 & 0.31 & 0.39 & 0.46 & 0.51 & 0.54 & 0.55 & 0.56 \\
CLIP-L-14 (Generated)\tnote{2}   & 0.02 & 0.32 & 0.43 & 0.50 & 0.57 & \textbf{0.60} & \textbf{0.61} & \textbf{0.62} \\
\hline
\end{tabular}
\begin{tablenotes}
\item[1] \textit{Original}: Images generated from the 3D point clouds by the Depth-Aware Image Generation algorithm without smoothing (Step 4).
\item[2] \textit{Generated}: Images generated by the proposed Depth-Aware Image Generation algorithm.
\end{tablenotes}
\end{threeparttable}
\end{table*}

\section{Results and Discussion}
\label{sec:results}

\subsection{Comparative Analysis of Few-Shot Learning Performance}
To evaluate the proposed framework, we conducted a comparative analysis using VLMs across varying model architectures and shot counts. Table \ref{tab:fewshot-results} summarizes the performance trajectories of CLIP and EVA-L models on the 20-class vehicle dataset. The results yield three main observations regarding the interplay between model scale, data preprocessing, and few-shot adaptation.

\begin{figure}[!t]
\centering
\subfloat[Original 2D Projection without depth-aware smoothing\label{fig:depth_original}]{
  \includegraphics[width=1.1\linewidth]{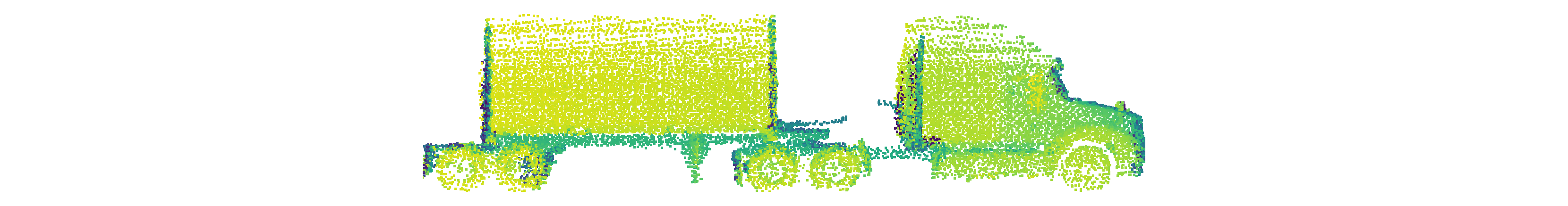}
}
\hfill
\subfloat[Depth-aware Generated\label{fig:depth_opening}]{
  \includegraphics[width=01.1\linewidth]{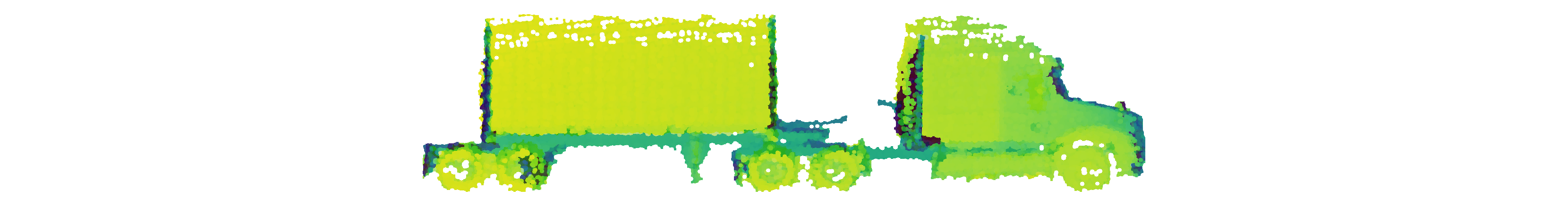}
}
\caption{Illustration of the 2D projected image without depth-aware smoothing (Step 4 of Algorithm 1) and its processed version after applying the entire Algorithm 1.}
\label{fig:depth_comparison}
\end{figure}

\subsubsection{Impact of Depth-aware Smoothing in the Image Generation Algorithm}
A key finding of this study is the architecture-dependent impact of the image generation pipeline proposed in the proposed Depth-aware Image Generation algorithm. To inspect the impacts of the depth-aware smoothing, {\it i.e.,} Step 4 of Algorithm 1, we performed tests on the {\it Generated} images, which are 2D images generated by calling Algorithm 1 in its entirety, and {\it Original} images, which are the 2D images generated by calling Algorithm 1 without Step 4. The sample results of the {\it Original} and {\it Generated} images for one sample 3D point cloud are depicted in Fig. \ref{fig:depth_comparison}, where it can be observed that the depth-aware smoothing invoked by Step 4 of Algorithm 1 indeed yields smoother images. As reported in Table III, for the larger \textbf{CLIP-ViT-L/14} model, the depth-aware smoothing operation yields a higher performance gain, elevating the 30-shot F1-score from 0.56 to 0.62. We attribute this to the fact that ViT-L/14, trained on large datasets of high-resolution natural images, tends to rely on coherent structural shapes. The proposed depth-aware smoothing operations in Step 4 of Algorithm 1 fill gaps in the sparse LiDAR projection, creating a denser visual proxy that aligns better with the model's pre-trained representations.

Conversely, the smaller \textbf{ViT-B/32} model shows slightly worse performance with processed images, dropping from 0.61 to 0.56 at 30 shots. This suggests that smaller architectures may rely more on high-frequency, point-level textures—artifacts that are smoothed out during morphological reconstruction. This observation highlights a design trade-off: morphological smoothing enhances structural coherence for high-capacity models but may remove micro-features utilized by more localized encoders.

\subsubsection{Attention Analysis: Original vs. Depth-aware generated images}
\begin{figure*}[htbp]
\centering
\includegraphics[width=\textwidth]{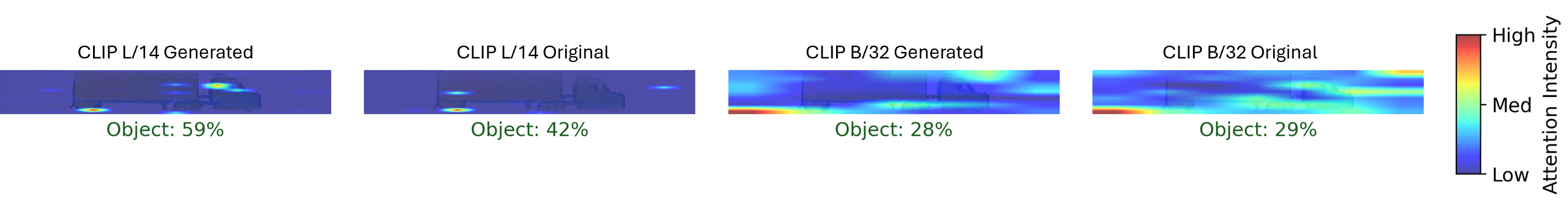}
\caption{Attention map comparison: CLIP-L/14 attention on the original 2D LiDAR projection (left) versus the depth-aware generated image (right). The proposed depth-aware smoothing yields more focused attention on the vehicle structure, whereas the original projection without depth-aware smoothing elicits diffuse attention on sparse artifacts.}
\label{fig:attention_opening_vs_original}
\end{figure*}

To interpret how depth-aware smoothing operations used in Step 4 of Algorithm 1 benefit the CLIP-L/14 encoder, we visualize the attention maps of CLIP's vision transformer processing both the {\it original} 2D projection (cf. Table III) and the depth-aware generated image (Fig. \ref{fig:attention_opening_vs_original}). On the original projection without depth-aware smoothing, attention appears diffuse and scattered across sparse point artifacts and background regions, indicating a struggle to extract coherent structure from noisy LiDAR data. In contrast, the depth-aware generated images elicit focused attention on the vehicle silhouette—specifically the cab, trailer, and structural contours. This shift from diffuse to structure-focused attention correlates with the observed performance gain for CLIP-L/14 (F1: 0.56 $\rightarrow$ 0.62). Furthermore, we observe a scaling effect: the larger CLIP-L/14 model demonstrates a significantly higher concentration of attention on object details compared with the smaller CLIP-B/32.

\subsubsection{Few-Shot Scaling and Stability}
All models show consistent improvements as the shot count increases. In the ultra-low-shot regime ($k=1$ to $4$), we observed higher variance ($\sigma \approx 0.05$), indicating that classification is sensitive to the choice of prototypes. Performance stabilizes beyond 16 shots ($\sigma \leq 0.02$). This convergence suggests that for real-world deployment, a support set of approximately 20--30 examples may be sufficient to construct a robust decision boundary, substantially reducing data requirements compared to supervised baselines that typically require thousands of samples to avoid overfitting \cite{li2021truck}.

\subsection{Fine-Grained Classification Analysis}
To assess the model's ability to distinguish subtle inter-class differences, we analyze the confusion matrix for the best-performing configuration (CLIP-L/14, 30-Shot).

\begin{figure*}[htbp]
\centering
\includegraphics[width=\textwidth]{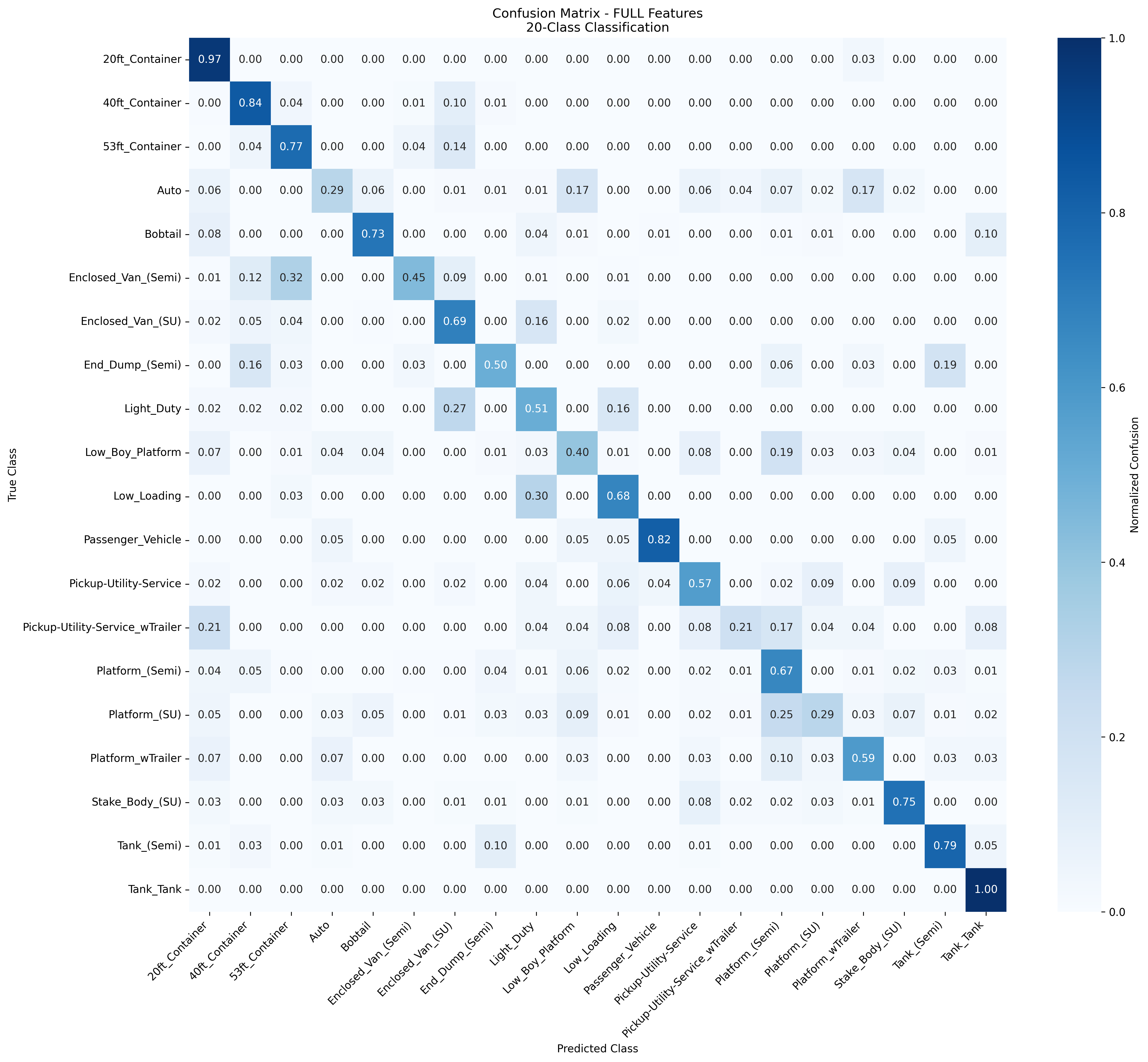}
\caption{Confusion matrix of the 20-class vehicle classification system using CLIP-L/14 (30-shot). Diagonal dominance indicates high classification accuracy; off-diagonal elements reveal geometric ambiguities between container classes.}
\label{fig:confusion_matrix}
\end{figure*}

Figure \ref{fig:confusion_matrix} shows strong diagonal dominance, indicating effective classification overall. However, specific error modes highlight the ``Half-Shell'' challenge inherent to roadside LiDAR:
\begin{itemize}
    \item \textbf{High Accuracy:} Distinct categories such as \textit{Tank Tank} and \textit{20ft Container Trucks} achieve high classification accuracy due to their unique silhouettes.
    \item \textbf{Geometric Ambiguity:} Misclassifications are concentrated among structurally similar classes, specifically \textit{53ft Containers} and \textit{Enclosed Van (Semi)}. From a side-fire LiDAR perspective, these vehicles share similar vertical profiles and are distinguished primarily by length. When occlusion occurs or the scan is partial, the geometric cues for length may be compromised, leading to inter-class confusion.
\end{itemize}

\subsection{Mechanistic Interpretability: Visual vs. Text Features}
A contribution of this work is analyzing the optimal modality for prompting VLMs in specialized engineering domains. We systematically analyzed the contribution of text versus visual embeddings.

\subsubsection{Analysis Setup: Prototype Fusion}
% To quantify the impact of semantic information, we modeled the classification prototype as a weighted linear combination of visual and textual embeddings. The fused prototype $p_c^{\text{fused}}$ for class $c$ is defined as:

% \begin{equation}
% \label{eq:fusion}
% p_c^{\text{fused}} = (1-w) \cdot p_c^{\text{visual}} + w \cdot e_c^{\text{text}}
% \end{equation}

% \noindent where $p_c^{\text{visual}}$ is the mean embedding of the visual support set, $e_c^{\text{text}}$ is the text embedding from the language encoder, and $w \in [0, 1]$ represents the scalar text weight. 

We evaluated the classification performance by visual and textual information as dictated by Eq. (\ref{eq:fusion}) by varying $w$ from $0.0$ (Pure Visual) to $1.0$ (Pure Text).

\begin{figure}[htbp]
\centering
\includegraphics[width=\columnwidth]{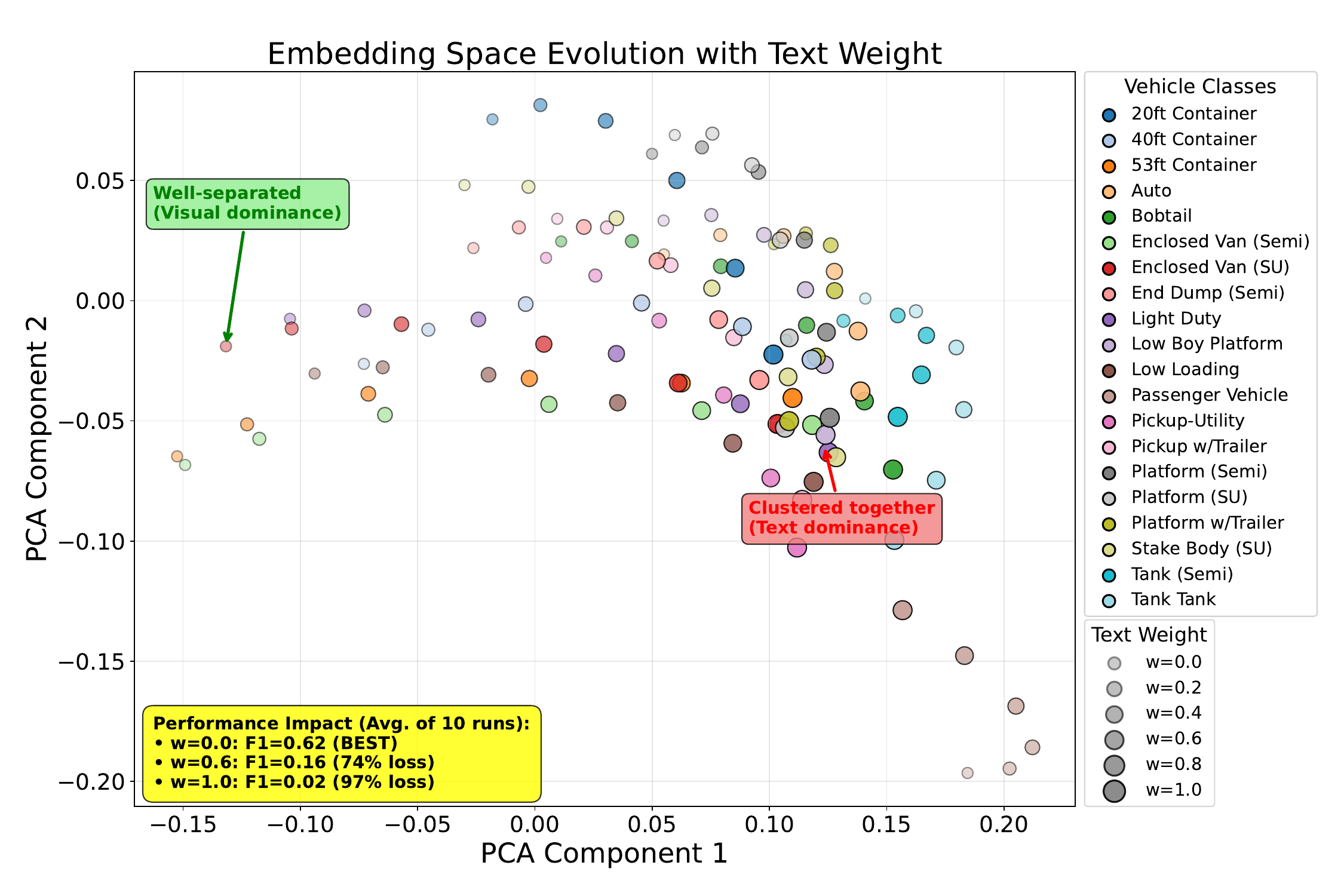}
\caption{Embedding space evolution with increasing text weight ($w$) in CLIP-based classification. Visual embeddings ($w=0$) show well-separated class clusters; text embeddings ($w=1$) collapse into high inter-class similarity. While heavy text weighting degrades performance overall, a minimal weight ($w=0.2$) provides useful regularization in low-shot regimes.}
\label{fig:embedding_space_evolution}
\end{figure}

\subsubsection{The Semantic Gap and Low-Shot Regularization}
Figure \ref{fig:embedding_space_evolution} illustrates the evolution of the embedding space as the weight of the text encoder ($w$) increases. The analysis reveals a nuanced trade-off between semantic priors and visual evidence:

\textbf{1. The Domain Gap (High $w$):} As $w$ approaches $1.0$ (Pure Text), performance drops to near-random levels (F1=0.019), representing a 96.2\% degradation. This provides empirical evidence of a \textbf{Semantic Gap}: CLIP's text encoder is aligned with natural language descriptions from the web, not the technical taxonomy of transportation engineering. Text embeddings for fine-grained classes share high cosine similarity ($>0.89$), limiting discriminative power.

\begin{figure}[htbp]
\centering
\includegraphics[width=\columnwidth]{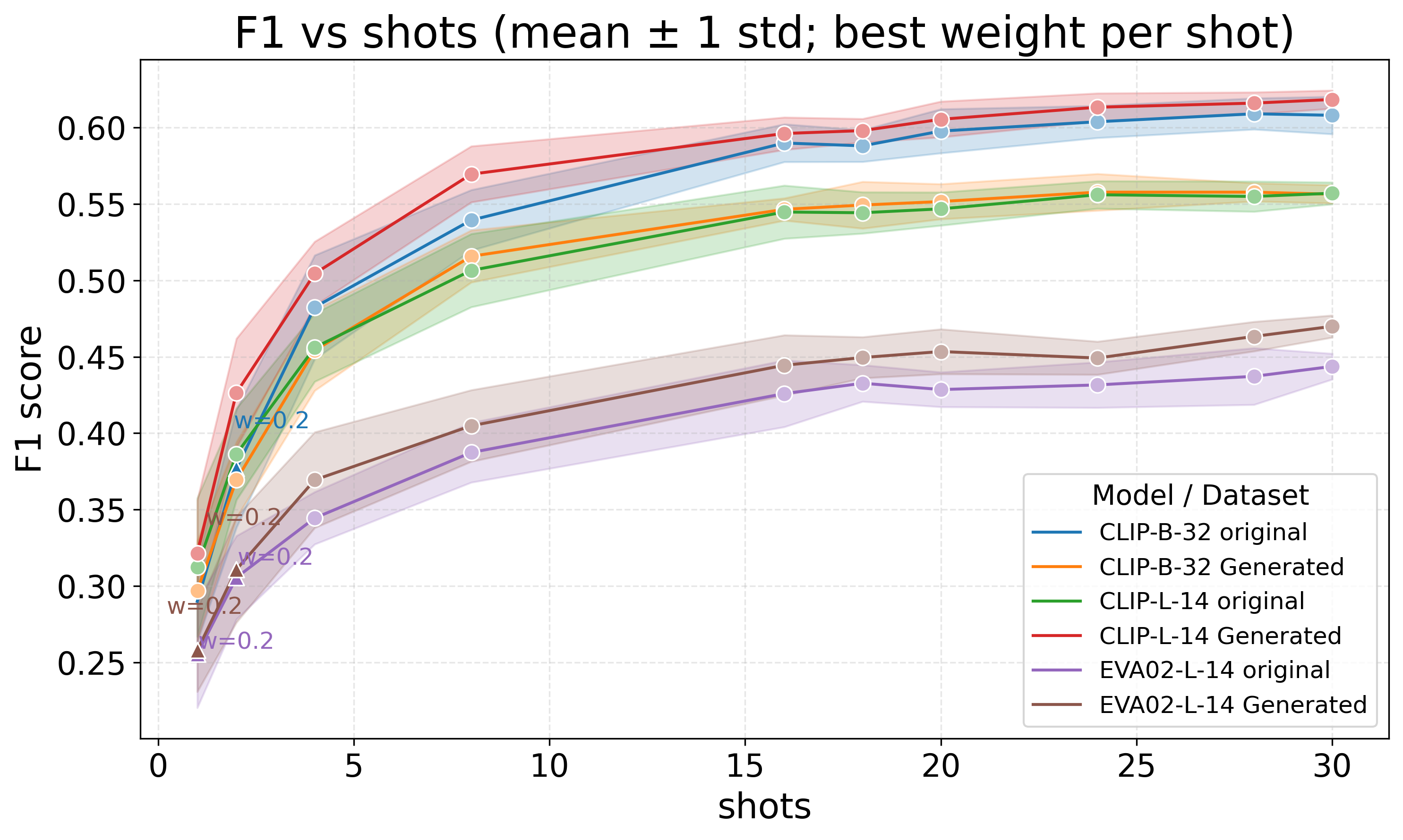}
\caption{Few-Shot Learning Results illustrating the performance trajectory across shot counts ($k=1$ to $30$). Triangle markers ($w=0.2$) outperform round markers ($w=0$) in few-shot regimes ($k<4$), indicating that semantic priors provide stability when visual samples are scarce. As data increases, pure visual prototypes ($w=0$) become dominant. ViT-B/32 and ViT-L/14 denote the CLIP ViT-Base (patch size 32) and CLIP ViT-Large (patch size 14) architectures, respectively.}
\label{fig:fsl}
\end{figure}

\textbf{2. The ``Semantic Anchor'' Effect (Low $w$):} Despite the domain gap, Figure \ref{fig:fsl} reveals an exception in the low-shot regime ($k \le 4$). Here, a slight text bias ($w=0.2$, triangle markers) consistently outperforms pure visual prototypes ($w=0$). This suggests that when visual examples are scarce, the semantic prior from text can act as a stabilizing ``anchor,'' regularizing noisy visual features. 

\textbf{3. Visual Dominance:} As the number of shots increases ($k > 8$), the visual evidence becomes sufficiently robust to define class boundaries on its own. At this stage, the ambiguous semantic information from the text encoder becomes detrimental, explaining why the pure visual model ($w=0$) achieves the best performance (F1=0.62) at 30 shots. For the comparative analysis in Section \ref{sec:ablation}, we use pure visual prototypes ($w=0$) to reflect this configuration. This transition suggests a characteristic of industrial few-shot learning: \textit{semantic priors can help for a cold start, but visual data becomes more important as data increases.} 

% \textcolor{red}{[ref] or discovery/observation by this work?} This is our findings

% \subsection{Attention Pattern Analysis}
\subsection{Training-Free vs. Supervised Adaptation}
\label{sec:ablation}

A key choice in few-shot deployment is between training-free adaptation and supervised fine-tuning. We evaluated our few-shot learning approach against a Linear Probe (CLIP-L/14) and fully supervised baselines. The Linear Probe also uses our Depth-Aware Morphological Reconstruction pipeline for input generation, ensuring a controlled comparison where input data quality is identical and the impact of the adaptation strategy is isolated.

\subsubsection{Performance and Cost Comparison}
Table \ref{tab:multishot_time_cost} presents a comparative analysis at $k=30$ shots per class. The Linear Probe (80 epochs) achieves a slightly higher F1-score (0.649) than the Training-Free Few-Shot method (0.621). This close proximity suggests that our reconstruction pipeline generates highly discriminative feature representations, as even a simple linear classifier can separate classes effectively. Furthermore, both VLM-based methods substantially outperform the supervised CLIP-B/16 baseline (0.087), which entirely fails to learn meaningful features from scratch given the limited data. The PointNet baseline ($F1=0.705$, 4.18 ms) serves as a specialized 3D reference. The high performance achieved with only $k=30$ examples validates our proposed “Label-and-Distill” (Cold Start) strategy: since 30 examples are sufficient to train a lightweight 3D network to high performance, our VLM framework can be used to automatically generate these labels, enabling the deployment of efficient PointNet models without manual annotation. Most notably, the Training-Free Few-Shot method offers a superior adaptation strategy: Unlike supervised methods that require iterative optimization and hyperparameter tuning, the training-free approach adapts in just 11.4 seconds—requiring only a single forward pass for feature extraction. This makes it an extremely agile solution for rapid deployment for practical use, while the image reconstruction latency remains an acceptable offline cost within an auto-labeling workflow.

\begin{table*}[t]
\centering
\caption{Comparison of methods at $k=30$ shots. The 2D methods use our Depth-Aware Reconstruction pipeline; PointNet uses the intermediate fused 3D reconstruction.}
\label{tab:multishot_time_cost}
\setlength{\tabcolsep}{5pt}
\renewcommand{\arraystretch}{1.2}
\begin{tabular}{l l c c c c}
\toprule
\textbf{Method} & \textbf{Input Representation} & \textbf{Training?} & \textbf{F1 @ 30} & \textbf{Adapt. Time (s)} & \textbf{Infer. Time (ms)} \\
\midrule
\multicolumn{6}{l}{\textit{Supervised Baselines}} \\
ViT-B/16 & Proposed 2D Proxy & Yes & 0.087 $\pm$ 0.012 & 300.4 & 8.97 $\pm$ 0.32 \\
PointNet & Fused 3D Point Cloud & Yes & \textbf{0.705} $\pm$ 0.142 & 198.3 & \textbf{4.18} $\pm$ 0.21 \\
\midrule
\multicolumn{6}{l}{\textit{VLM Adaptation Strategies}} \\
\textbf{Ours (Linear Probe)} & \textbf{Proposed 2D Proxy} & Yes & 0.649 $\pm$ 0.008 & 12.4 & 19.54 $\pm$ 0.30 \\
\textbf{Ours (Few-Shot Learning)} & \textbf{Proposed 2D Proxy} & \textbf{No} & 0.621 $\pm$ 0.007 & \textbf{11.4} & 19.76 $\pm$ 0.25 \\
\bottomrule
\end{tabular}
\end{table*}

\subsubsection{The ``Training Trap'' in Data-Scarce Regimes}
\begin{figure}[htbp]
\centering
\includegraphics[width=\columnwidth]{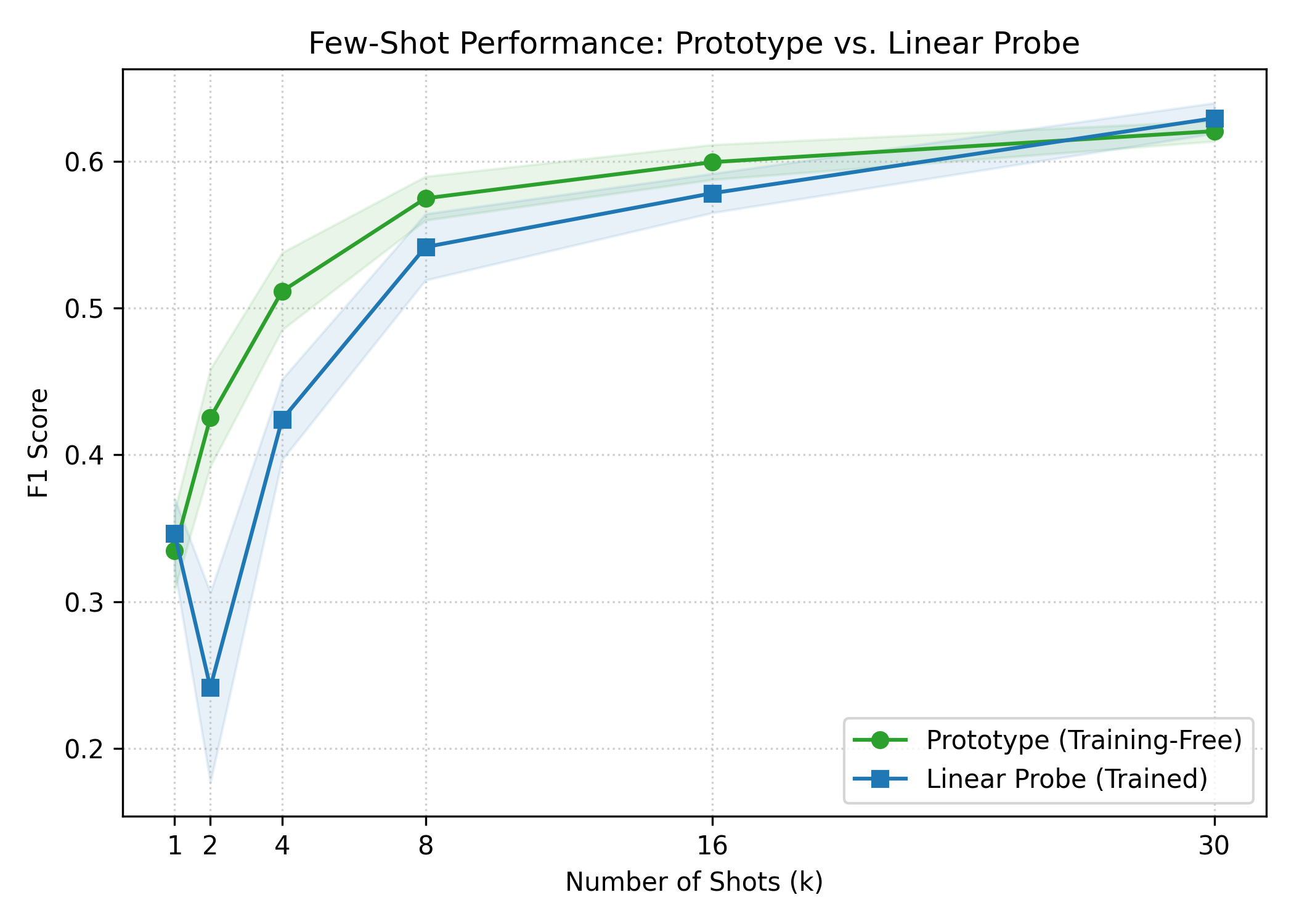}
\caption{Data efficiency comparison: The training-free few-shot learning method (green) outperforms the Linear Probe (blue) in low-data regimes ($k < 16$). The Linear Probe exhibits a ``Training Trap'' at $k=2$, where performance drops due to overfitting.}
\label{fig:comparison_shots_full}
\end{figure}

While the Linear Probe shows a slight advantage at $k=30$, our analysis indicates that it is unstable in true few-shot regimes. Figure \ref{fig:comparison_shots_full} illustrates this ``Training Trap.'' At $k=2$ shots, the Linear Probe's performance drops to F1=0.24 due to overfitting, whereas the few-shot learning method maintains an acceptable F1 score of 0.43. 

The training-free few-shot learning approach is statistically significantly superior ($p < 0.001$, paired t-test) for all low-data settings ($k \in \{2, 4, 8, 16\}$). It exhibits a monotonic performance curve, yielding predictable behavior with minimal supervision. In contrast, the Linear Probe becomes competitive only when the data volume is sufficient to constrain optimization of the classification head ($k \ge 30$), which could be risky in practical utilities.

\subsubsection{Sensitivity and the Validation Paradox}
\begin{figure}[htbp]
\centering
\includegraphics[width=\columnwidth]{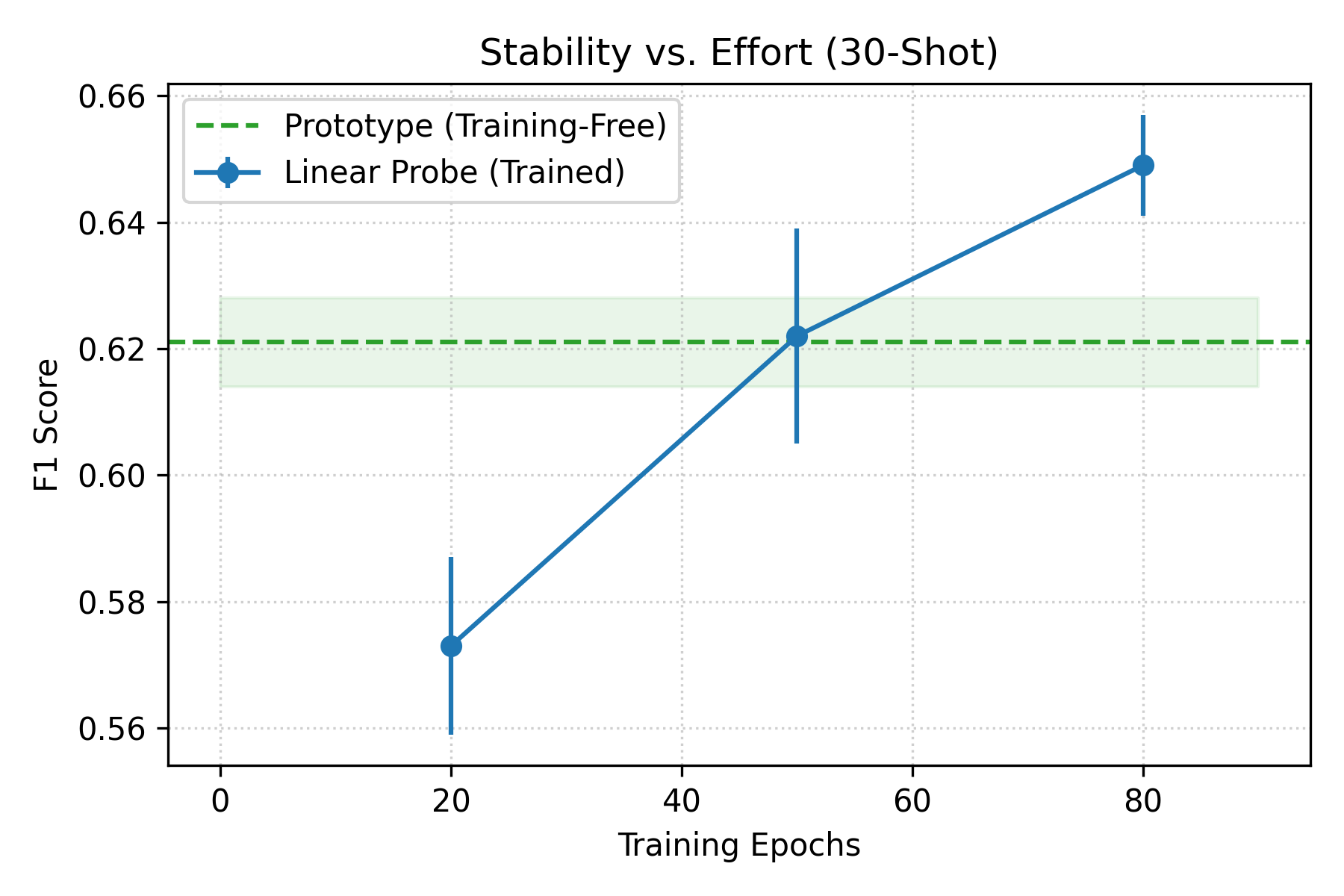}
\caption{Sensitivity analysis (30-shot): Linear Probe performance varies with training epochs. Selecting a suboptimal stopping point (e.g., 20 epochs) yields performance below the training-free few-shot learning baseline. The few-shot learning method provides stable, hyperparameter-free performance.}
\label{fig:sensitivity_epochs}
\end{figure}

The sensitivity analysis in Figure \ref{fig:sensitivity_epochs} further supports the training-free approach. Achieving F1=0.649 with the Linear Probe required training for 80 epochs. An undertrained model (20 epochs) yielded F1=0.57, underperforming the parameter-free few-shot learning (0.621).

This highlights a ``Validation Paradox'' in few-shot learning: determining optimal hyperparameters (e.g., 80 epochs) typically requires a validation set. With only 30 labeled examples, setting aside data for validation is costly. The few-shot learning method, with no hyperparameters to tune, avoids this issue and offers consistent baseline performance without validation data, which is more practical in real-world applications.

\section{Conclusion and Future Work}
\label{sec:conclusion}

This study presents a framework that bridges the modality gap between infrastructure-based LiDAR systems and VLMs, addressing the scalability bottleneck in intelligent transportation systems. By introducing a \textbf{Depth-Aware Morphological Reconstruction} pipeline, we transform sparse, occluded point clouds into visual proxies that align with the pre-trained feature space of models like CLIP. To our knowledge, few studies have systematically validated the transferability of VLM priors to the domain of roadside LiDAR classification.

Our experimental analysis yields three main contributions:
\begin{enumerate}
    \item \textbf{Operational Scalability \& Auto-Labeling}: The proposed framework achieves robust classification accuracy with as few as 16–30 examples per class. We suggested a “VLM-Guided Supervision” workflow: utilizing the training-free VLM to bootstrap a dataset enables the subsequent training of efficient 3D supervised models (e.g., PointNet). The supervised student model achieves higher accuracy ($0.705$ vs. $0.621$) and lower latency ($4$ ms vs. $20$ ms) compared to the VLM teacher, offering a practical path for real-time edge deployment. Furthermore, with larger datasets, such supervised models scale effectively which is able to achieve higher accuracy\cite{li2021truck}.
    
    \item \textbf{The ``Semantic Anchor'' Phenomenon:} We observe a trade-off between visual and textual modalities. While text embeddings generally degrade performance due to the domain gap between technical taxonomy and natural language, they provide useful regularization in ultra-low-shot regimes ($k <  4$). This suggests that semantic priors can act as a stabilizing anchor when visual evidence is scarce, while visual-centric adaptation tends to be preferable as data increases.
    
    \item \textbf{Robustness in Data-Scarce Regimes:} Our comparative analysis indicates that supervised linear probing (using the same depth-aware inputs) can marginally outperform training-free methods given sufficient data ($k \ge 30$), but it exhibits marked performance degradation in true few-shot settings ($k \le 4$). In contrast, our few-shot learning approach demonstrates monotonic stability and superior data efficiency, making it particularly well-suited for ``cold start'' deployment scenarios where initial data is scarce.
\end{enumerate}

To address the remaining challenge of fine-grained geometric ambiguity (e.g., distinguishing semi-trailer Enclosed Vans from 53ft containers), our future work will focus on Grounded Spatial Reasoning. We plan to conduct a quantitative analysis of attention dynamics, moving beyond qualitative visualization to compute metrics such as \textit{Attention Mass Concentration} and \textit{Region-of-Interest (RoI) Overlap}. By correlating these metrics with misclassification cases, we can pinpoint exactly where holistic embeddings fail to attend to discriminative features. Guided by these insights, we aim to integrate grounded vision models (e.g., Florence-2 or Grounding DINO) to explicitly isolate and measure specific vehicle subsections—such as trailer overhangs or axle spacing—thereby resolving the subtle structural differences that currently challenge pure classification approaches.

\newpage

\bibliographystyle{IEEEtran}
\bibliography{reference}
\end{document}